\documentclass[copyright,creativecommons]{eptcs}

\providecommand{\event}{Post Proceedings of ThEdu'19}

\usepackage[english]{babel}
\usepackage[utf8x]{inputenc}
\usepackage[T1]{fontenc}
\usepackage{url}
\usepackage{hyperref}
\usepackage{breakurl}

\def\GASC{\emph{GASC}}

\title{Towards a Geometry Automated Provers Competition\thanks{This
    work is financed by national funds through the FCT - Foundation
    for Science and Technology, I.P., within the scope of the project
    CISUC - UID/CEC/00326/2019}}

\author{Nuno Baeta
   \institute{
     CISUC,\\
     University of Coimbra, Portugal}
  \email{nmsbaeta@gmail.com}
  \and
  Pedro Quaresma
  \institute{
    CISUC/Department of Mathematics,\\
    University of Coimbra, Portugal}
  \email{pedro@mat.uc.pt}
  \and
  Zolt\'an Kov\'acs
  \institute{
    The Private University College of Education of the Diocese of Linz\\
    Linz, Austria}
  \email{zoltan@geogebra.org} \\
}

\def\authorrunning{N. Baeta, P. Quaresma \& Z. Kov\'acs}

\def\titlerunning{Towards a Geometry Automated Provers Competition}

\def\copyrightholders{N. Baeta, P. Quaresma \& Z. Kov\'acs}

\begin{document}

\maketitle

\begin{abstract}
  The geometry automated theorem proving area distinguishes itself by
  a large number of specific methods and implementations, different
  approaches (synthetic, algebraic, semi-synthetic) and different
  goals and applications (from research in the area of artificial
  intelligence to applications in education).

  Apart from the usual measures of efficiency (e.g.~CPU time), the
  possibility of visual and/or readable proofs is also an expected
  output against which the geometry automated theorem provers (GATP)
  should be measured.

  The implementation of a competition between GATP would allow to
  create a test bench for GATP developers to improve the existing
  ones and to propose new ones. It would also allow to establish a
  ranking for GATP that could be used by ``clients'' (e.g.~developers
  of educational e-learning systems) to choose the best implementation
  for a given intended use.
\end{abstract}


\setcounter{tocdepth}{2}

%
%


\section{Introduction}
\label{sect:introduction}

The area of geometry automated theorem proving distinguishes itself by
a large number of specific methods and implementations. Synthetic
methods try to automate the traditional geometric proving
processes~\cite{Stojanovic2011,Wu1984}; although being able to produce
readable proofs, the so far proposed methods are very narrow-scoped
and not efficient. The algebraic methods reduce the complexity of
logical inferences by translating the geometric conjecture to an
algebraic conjecture and then applying a given algebraic method. What
is gained in efficiency and wider scope is lost in the connection of
the algebraic proof and the geometric reasoning. These methods are
broad-scope and efficient. However, if eventually a proof record is
produced, it will be a very complex algebraic proof~\cite{Wu1984}. In
order to combine the geometric reasoning of synthetic methods and the
efficiency of algebraic methods, some approaches, such as the area
method and the the full-angle method, represent geometric knowledge in
a form of expressions with respect to geometric invariants. These
methods are broad-scoped, efficient and capable of producing geometric
proofs~\cite{Chou1996a,Chou1996b,Janicic2012a}.


When considering the geometric automated theorem provers (GATP),
questions of applicability, e.g.~in education, are very important. The
improvement of existing implementations or the goals to be attained by
new methods/implementations must take in consideration not only
research goals but also the practical intended usefulness.

To be able to compare the different methods and implementations, a
competition will have the virtue of pushing towards the
standardization of the input language, the standardization of test
sets, the direct comparability and the easier exchange of ideas and
algorithmic techniques. The results of such a competition will also
constitute a showcase, where potential users will look for the best
GATP for their goals.

Towards a \emph{Geometry Automated theorem provers System Competition}
(\GASC) many steps must be develop and combined. Quoting from
TOOLympics
2019\footnote{\url{https://tacas.info/toolympics.php}}~\cite{Bartocci2019,toolympics2019}:

\begin{enumerate}
\item How to assess adequacy of benchmark sets, and how to establish
  suitable input formats? And what is a suitable license for a
  benchmark collection?

\item How to execute the challenges (on-site vs. off-site, on
  controlled resources vs. on individual hardware, automatic
  vs. interactive, etc.)?

\item How to evaluate the results, e.g. in order to obtain a ranking?

\item How to ensure fairness in the evaluation, e.g. how to avoid
  bias in the benchmark sets, how to reliably measure execution times,
  and how to handle incorrect or incomplete results?
  
\item How to guarantee reproducibility of the results?
  
\item How to achieve and measure progress of the state of the art? 

\item How to make the results and competing tools available so that
  they can be leveraged in subsequent events? 
\end{enumerate}

Some partial results are already available: a common language to state
the geometric theorems~\cite{Quaresma2015a}, a comprehensive
repository of geometric problems~\cite{Quaresma2011} and a set of
measures of quality capable of assessing the GATPs in different
classes~\cite{Baeta2019,Quaresma2019}.

The ideas behind such a competition were presented
in~\cite{Baeta2019}. From the subsequent discussion, where a small set
of six geometric problems was chosen, we progressed to an actual
competition at ThEdu'19, GASC 0.1, run in a local computer, and where
a set of GATPs competed over problems in the TGTP database.  This
paper is the result of all the discussions occurred there. For GASC
0.2 the complete set of problems in TGTP was used and the competition
was conducted over the Internet, using a Web-server to run the
competition and clients to check for the running of the competition
and its final results.

\medskip
\textit{Overview of the paper.}  The paper is organised as follows:
first, in Section~\ref{sec:gatps}, some current GATP are presented.
In Section~\ref{sec:tgtp} the repository of geometric problems is
presented. In Section~\ref{sec:i2gatp}, the question related to the
common format are discussed. In Section~\ref{sec:competition} the
practical question about the implementation of the competition are
discussed. In Section~\ref{sec:results} the different measures of
quality are discussed. Final conclusions are drawn and future work is
foreseen in Section~\ref{sec:futurework}.


\section{GATPs}
\label{sec:gatps}

For the preliminary run of {\GASC}, during
ThEdu'19,\footnote{\href{http://www.uc.pt/en/congressos/thedu/thedu19}{ThEdu'19},
  the 8th International Workshop on Theorem proving components for
  Educational software, 25 August 2019, Natal, Brazil} the \emph{GCLC}
set of provers and some of the provers in \emph{GeoGebra's portfolio
  prover} were
selected~\cite{Botana2015,Janicic2012a,Kovacs2015}. The
criteria used was: availability, reliability, and the possibility of
running them in the command line (i.e.~in a stand-alone fashion,
outside a given computational tool where they could be integrated).

In future editions of {\GASC} we hope that such set of provers can be
enlarged as much as possible. A list of possible candidates is given
by: 

\begin{description}
\item[GCLC] \emph{GCLC} is a tool for visualizing objects and notions
  of geometry and other fields of mathematics, by generating figures
  and animations in the \emph{gc} language. It has a a built-in
  geometry theorem prover that can automatically prove a range of
  complex problems. The GATP module implements the \emph{Area Method},
  the \emph{Wu's method} and the \emph{Gr{\"o}bner Basis
    Method}~\cite{Janicic2006c,Janicic2012a}.

  The implemented GATP can be called from inside the
  \emph{gclc-workbench} (Linux) or \emph{WinGCLC} (MS-Windows) DGS
  tools, but can be also used in an independent way. Whenever called,
  in the command line it will produce (if successful) a rendering of
  the construction and a proof record, both to be processed by a
  {\LaTeX} compiler.

  \begin{itemize}
  \item[] \texttt{\$ gclc GEO0001.gcl}\hfill Area Method (default
    method)~\cite{Janicic2012a}; 
  \item[] \texttt{\$ gclc GEO0001.gcl -w} \hfill Wu's method~\cite{Chou1994};
  \item[] \texttt{\$ gclc GEO0001.gcl -g} \hfill Gr{\"o}bner bases method~\cite{Chou1994}.
  \end{itemize}
\item[OpenGeoProver] Open Library of Geometry Automatic Theorem
  Provers,
  \emph{OpenGeoProver}.\footnote{\url{https://github.com/opengeometryprover}}
  It is an open source project, aiming to implement various geometry
  automated theorem provers.  It can be used as a stand-alone tool but
  can also be integrated into other geometry tools, such as dynamic
  geometry software, e.g. work is being made to integrate
  \emph{OpenGeoProver} with GeoGebra~\cite{Petrovic2012a}.
  In its current state, \emph{OpenGeoProver} implements the Wu's
  method. Some work has already been done to include implementations
  of the area method and the full-angle method~\cite{Baeta2013}.

  \begin{itemize}
  \item[] \texttt{\$ ./runOGP GEO0001.xml} \hfill Wu's
    method~\cite{Chou1994}; 
  \end{itemize}
  this is a \emph{bash} script that calls the OGP prover (\emph{Java}
  bytcode).
\item[CoqAM] The formalisation of the area method using the proof
  assistant \emph{Coq} was done by implementing the decision procedure
  as a \emph{Coq} tactic and formalising all theorems needed by the
  method. The implementation guarantee the soundness of the method
  implementation, i.e., the proofs generated by the tactic are always
  correct~\cite{Janicic2012a,Narboux2009}.
  
  \begin{itemize}
  \item[] \texttt{\$ coqc GEO0001.v > GEO0001.errors} \hfill Area
    Method in \emph{Coq}~\cite{Janicic2012a,Narboux2009}.
  \end{itemize}
\item[GeoGebra's portfolio prover] GeoGebra has an embedded prover
  system that is capable of using multiple internal backends for
  proving theorems~\cite{Kovacs2015}.  Its \texttt{Prove} and
  \texttt{ProveDetails} commands are the user level interface to
  formalize statements in the given syntax. They are considered as
  low-level commands because most users want to compare geometric
  objects directly by using the \textbf{Relation Tool} in GeoGebra,
  and by having an automated conjecture that is based on numerical
  checks, the low-level commands will also be issued by GeoGebra
  automatically.

  The backends include Recio's exact check
  method~\cite{recio-eaca-paper}, the Gr\"obner Basis Method, and
  \textit{OpenGeoProver} can also be internally used to perform
  computations via Wu's method. For the Gr\"obner Basis Method it is
  possible to use an internal implementation of computing Gr\"obner
  bases via the \textit{Giac} computer algebra
  system~\cite{GiacGG-RICAM2013}, or to use an external system that
  uses \textit{Singular}~\cite{SWS-AMAI2014,sws-eaca-paper}.

  \begin{itemize}
  \item[] \texttt{\$ xvfb-run geogebra --prover=engine:Recio
      GEO0001.ggb} \hfill Recio's exact check
    method~\cite{recio-eaca-paper};  
  \item [] \texttt{\$ xvfb-run geogebra --prover=engine:Botana
      GEO0001.ggb} \hfill Gr\"obner Basis
    method~\cite{SWS-AMAI2014,sws-eaca-paper}.
  \end{itemize}

  GeoGebra's portfolio prover system automatically decides which
  backend should be used, but currently the best results can be
  obtained with the Gr\"obner Basis Method via
  \textit{Giac}~\cite{amai-portfolio}. On the other hand, the
  selection of the backend can be fine-tuned by using command line
  options of GeoGebra in its desktop version. GeoGebra internally has
  a problem repository that is used for testing each backend on a
  daily basis---the results are published and updated at
  \href{https://prover-test.geogebra.org/job/GeoGebra-provertest/ws/test/scripts/benchmark/prover/html/all.html}{prover-test.geogebra.org}\footnote{\url{https://prover-test.geogebra.org/job/GeoGebra-provertest/ws/test/scripts/benchmark/prover/html/all.html}}
  by a Jenkins
  system\footnote{\url{https://github.com/jenkinsci/jenkins}}
  installed at that \href{https://prover-test.geogebra.org/}{server},
  providing a continuous checking of the results.
\end{description}

There are many other GATPs to be considered: \emph{ArgoCLP}, a
Coherent Logic Based Geometry Theorem Prover~\cite{Stojanovic2011},
\emph{GEOTHER}~\cite{Wang2004},
\emph{Gex}~\cite{Gao1999,Gao1998b,Gao1998a},
\emph{JGEX}~\cite{Ye2011}, \emph{MMP}~\cite{Gao2004}; different
formalizations in Coq: an automatic prover for projective
geometry~\cite{Braun2018}; Gröbner basis
method~\cite{Pottier1998,Pottier2008}; Buchberger's
algorithm~\cite{Gregoire2011}; Wu's method~\cite{Geneaux2011}. The
first-order generic theorem provers must also be considered.

The challenge is to be able to incorporate them in GASC along any
other GATP not listed above and/or any new system in current
development. 

%




\section{Repositories of Geometric Problems}
\label{sec:tgtp}

To test the GATP a test suite of problems must be created for each
edition of the competition. The repository of problems \emph{Thousands
  of Geometric problems for geometric Theorem Provers} (TGTP ) could
be used for such effect. TGTP is a Web-based repository of problems in
geometry, with a significant size.\footnote{v2.1.91---236 problems,
  \url{http://hilbert.mat.uc.pt/TGTP}} It also provides a supporting
library to allow the use of the repository by different
GATP~\cite{Quaresma2011}.

The set of problems in consideration should also consider different
axiom systems: neutral geometry; euclidean geometry; hyperbolic
geometry; projective geometry; etc. Different types of conjectures
should also be considered: constructive geometry; ruler and compass
construction problems, conjectures involving inequalities. 

After each edition of the GASC the (eventually) new problems would
increase the TGTP repository.

\section{GATPs Common Format}
\label{sec:i2gatp}

To be able to proceed, involving more GATPs, diversifying the axiom
systems and the type of conjectures, a common format must be
developed. 

The \textsc{I2GATP} format is an extension of the \textsc{I2G}
(Intergeo) common format aimed to support conjectures and proofs
produced by geometric automatic theorem provers. The goal in building
such a format is to provide a communication channel between different
tools from the field of geometry, allowing the linking of such tools,
as well as allowing the use of geometric knowledge kept in different
repositories~\cite{Quaresma2015a}. The TGTP repository and
accompanying library of filters support the
\textsc{I2GATP}.\footnote{Library of filters supporting the I2GATP
  common format \url{https://github.com/GeoTiles/libI2GATP}}

Having that (or other common format) filters from the common format to
the new GATPs in the competition must be implemented just before the
start of the competition.\footnote{From GeoThms to GeoGebra:
  \url{https://github.com/kovzol/GeoThms2ggb}.}

\section{GASC 0.2 Competition}
\label{sec:competition}

Using the TGTP database a set of 224 problems (geometric conjectures)
were selected. The GCLC code and the Coq area method code were used. A
filter from the GCLC code to GeoGebra code was implemented. In the
future all the problems should be in a common format, with filters for
all and each GATP in competition.

The first step toward a \textbf{G}eometry \textbf{A}utomated theorem
provers \textbf{S}ystems \textbf{C}ompetition (GASC) was given at
ThEdu'19, a
\href{http://www.uc.pt/en/congressos/thedu/thedu19/ficheiros/opaThEdu3}{presentation}
was made and a first trial, GASC 0.1, was conducted in a local
computer (the second author laptop) using two scripts: one to launch
the competition and follow it and another script to see the results in
a never ending loop.

After ThEdu'19, GASC 0.2 was run incorporating all the comments
received during the workshop, e.g.~the TOOLympics
reference~\cite{toolympics2019}. The major difference between GASC 0.1
and GASC 0.2 is in the use of an Web server to support the
competition.\footnote{\url{http://hilbert.mat.uc.pt/GASC/}}

The server that supported \href{http://hilbert.mat.uc.pt/GASC}{GASC
  0.2} was a Linux system, \texttt{Linux 4.9.0-2-amd64 \#1 SMP Debian
  4.9.18-1 (2017-03-30) x86\_64 GNU/Linux}. The desktop computer
motherboard is a \texttt{Intel(R) Core(TM) i7-4770 CPU @ 3.40GHz} with
16GiB of RAM.

The possibility of continuing to run GASC on a dedicated server or to
change to a specialized platform like StarExec~\cite{Stump2014} is an
open question.

\section{Results \& Taxonomies}
\label{sec:results}

Apart the simple measure of speed of execution (CPU times), GATP
should also be evaluated by other criteria, such as: readability of
the proof produced and usability, e.g.~in an educational setting.

How to measure the readability of proofs is still a research
problem~\cite{Baeta2019,Quaresma2019a,Quaresma2019}, Chou proposed a
way to measure how difficult a formal proof is (using the area
method)~\cite{Chou1994}, de Bruijn also proposed a coefficient, the
de~Bruijn factor, the quotient of the \emph{size} of corresponding
informal proof and the \emph{size} of the formal proof, could also be
used as a measure of readability~\cite{Wiedijk2000}. This
is close to a Turing test for proofs: if a human cannot distinguish
the proof generated automatically from a human proof, than it is
readable.


Up to now this issue was not addressed. A first simple binary criteria,
has a readable geometric proof or not, could be used to start.

The other criteria, the GATP usability in an educational setting, can
be analysed in two different ways. The formal validation of a given
conjecture, i.e.~given a construction done using a DGS the possibility
of having a formal validation of conjectures over that construction, or
the use of the proof as a learning object by itself.

For the validation of conjectures the important factor is (again) the
time, or more precisely the ``wait-time''---periods of silence that
followed teacher questions and students' completed
responses~\cite{Rowe1972,Stahl1994}. The following classes of ``GATP
validation time'' could be defined, in terms of time, $t$, taken by
the GATP to answer~\cite{Quaresma2019}:

\begin{itemize}
\item good: $t \leq 1.5s$;
\item fair: $1.5s < t \leq 3s$;
\item poor: $t > 3s$.
\end{itemize}

For the use of the proof as learning objects, we are back to the
readability of GATP proofs, so, again, a binary choice between
``\texttt{maybe}'' or ``\texttt{not available}'' is, for now, the only
possible outcome.

\section{Future Work}
\label{sec:futurework}

All the research and technical issues about the competition, described
above, must be solved/fixed. 

The organization of the competition in the long term would require the
support of the geometry automated deduction community: by entering the
competition; by setting a problems committee that would choose the set
of problems to be solved by the GATP, and maybe the more important
point, by using its outcomes to their research and/or applications.

It is planned that a new zero-edition (0.3) will be implemented at
ThEdu'20 (workshop at the International Joint Conference on Automated
Reasoning, IJCAR 2020, June 29--July 5, 2020, Paris, France) and that
the first edition of GASC would occur at the 13th International
Workshop on Automated Deduction in Geometry, ADG 2020, Hagenberg,
Austria, 13-15 July, 2020.

\bibliographystyle{eptcs}



\newcommand{\noopsort}[1]{}\newcommand{\singleletter}[1]{#1}

\end{document}